\documentclass{article}
\PassOptionsToPackage{numbers,sort}{natbib}
\usepackage[preprint]{neurips_2026}
\usepackage[utf8]{inputenc} 
\usepackage[T1]{fontenc}    
\usepackage{hyperref}       
\usepackage{url}            
\usepackage{array}
\usepackage{amsfonts}       
\usepackage{amsmath,amssymb}
\usepackage{mathtools}
\usepackage{cancel}         
\usepackage{nicefrac}       
\usepackage{xcolor}         
\usepackage{graphicx}
\usepackage{subcaption}
\usepackage{placeins} 
\usepackage[nameinlink]{cleveref}
\usepackage{float} 
\usepackage{enumitem}
\hypersetup{colorlinks, linkcolor={red!65!black}, citecolor={blue!75!black}, urlcolor={blue!75!black}} 

\usepackage{listings}
\usepackage{inconsolata}
\definecolor{codebg}{HTML}{F7F7F7}
\definecolor{codeframe}{HTML}{D8D8D8}
\definecolor{codekw}{HTML}{008000}
\definecolor{codestr}{HTML}{BA2121}
\definecolor{codecomment}{HTML}{3D7B7B}
\definecolor{codelineno}{HTML}{6A737D}
\lstdefinestyle{paperpython}{
  language=Python,
  basicstyle=\ttfamily\fontsize{9pt}{11pt}\selectfont,
  keywordstyle=\color{codekw}\bfseries,
  commentstyle=\color{codecomment}\itshape,
  stringstyle=\color{codestr},
  numbers=left,
  numberstyle=\scriptsize\color{codelineno},
  numbersep=8pt,
  frame=single,
  rulecolor=\color{codeframe},
  framesep=6pt,
  backgroundcolor=\color{codebg},
  showstringspaces=false,
  columns=fullflexible,
  keepspaces=true,
  tabsize=4,
  breaklines=false,
  upquote=true,
}

\definecolor{deep_blue}{HTML}{4c72b0}
\definecolor{deep_orange}{HTML}{dd8452}
\definecolor{deep_green}{HTML}{55a868}
\definecolor{deep_red}{HTML}{c44e52}
\definecolor{deep_purple}{HTML}{8172b3}

\definecolor{gentlelilac}{RGB}{112, 48, 175}
\definecolor{notegray}{RGB}{95, 95, 95}
\newcommand{\gentlelilaccancel}[1]{{\renewcommand{\CancelColor}{\color{gentlelilac}}\cancel{#1}}}

\newcommand{\viridisEight}[8]{%
  \textcolor[HTML]{440154}{#1}%
  \textcolor[HTML]{46327E}{#2}%
  \textcolor[HTML]{365C8D}{#3}%
  \textcolor[HTML]{277F8E}{#4}%
  \textcolor[HTML]{1FA187}{#5}%
  \textcolor[HTML]{4AC16D}{#6}%
  \textcolor[HTML]{A0DA39}{#7}%
  \textcolor[HTML]{FDE725}{#8}%
}

\newcommand{\plasmaEleven}[9]{%
  \textcolor[HTML]{F89540}{#1}%
  \textcolor[HTML]{F0804D}{#2}%
  \textcolor[HTML]{E3685F}{#3}%
  \textcolor[HTML]{D25073}{#4}%
  \textcolor[HTML]{BD3786}{#5}%
  \textcolor[HTML]{A72199}{#6}%
  \textcolor[HTML]{8F0DA4}{#7}%
  \textcolor[HTML]{7201A8}{#8}%
  \textcolor[HTML]{5302A3}{#9}%
  \plasmaElevenRest
}
\newcommand{\plasmaElevenRest}[2]{%
  \textcolor[HTML]{330597}{#1}%
  \textcolor[HTML]{0D0887}{#2}%
}

\newcommand{\tok}[1]{%
  \begingroup
  \setlength{\fboxsep}{1.2pt}%
  \colorbox{black!8}{\strut\texttt{#1}}%
  \endgroup
}

\newcommand{\toksep}{\allowbreak\hspace{0.12em}}

\usepackage{tikz}
\definecolor{screenshotgreen}{RGB}{95,170,100}
\definecolor{screenshotred}{RGB}{200,95,95}
\newcommand{\greenpoint}{%
  \tikz[baseline=-0.6ex]{
    \fill[screenshotgreen, opacity=0.55] (0,0) circle (0.8ex);
    \draw[screenshotgreen, opacity=0.9, line width=0.2ex] (0,0) circle (0.6ex);
  }%
}
\newcommand{\redcross}{%
  \tikz[baseline=-0.6ex]{
    \begin{scope}[rotate=45]
      \fill[screenshotred] (-1ex,-0.2ex) rectangle (1ex,0.2ex);
      \fill[screenshotred] (-0.2ex,-1ex) rectangle (0.2ex,1ex);
    \end{scope}
  }%
}

\title{Forgetting in Language Models: \\ Capacity, Optimization, and Self-Generated Replay}

\author{%
  Martin Marek \\
  New York University \\
  \texttt{martin.m@nyu.edu}
  \And
  Dongkyu Cho \\
  New York University
  \And
  Shikai Qiu \\
  New York University
  \AND
  Rumi Chunara \\
  New York University
  \And
  Pavel Izmailov \\
  New York University
  \And
  Andrew Gordon Wilson \\
  New York University
}

\begin{document}
\raggedbottom

\maketitle

\begin{abstract}
Models trained on a new task typically degrade on prior tasks, a phenomenon known as forgetting. Traditionally, mitigating forgetting has required replaying stored exemplars from prior tasks, which is often impractical. By contrast, language models can sample from their own training distribution, and we show that these self-generated samples serve as effective replay data, nearly eliminating forgetting. We find that forgetting nonetheless persists when the model has little remaining capacity: models pretrained close to saturation cannot absorb new information without overwriting prior knowledge. When capacity is not the limiting factor, low learning rates reduce forgetting but require substantially more training steps. Replay breaks this tradeoff, enabling fast, high-learning-rate finetuning without forgetting.
\end{abstract}

\begin{figure}[!h]
    \centering
    \includegraphics[width= \textwidth]{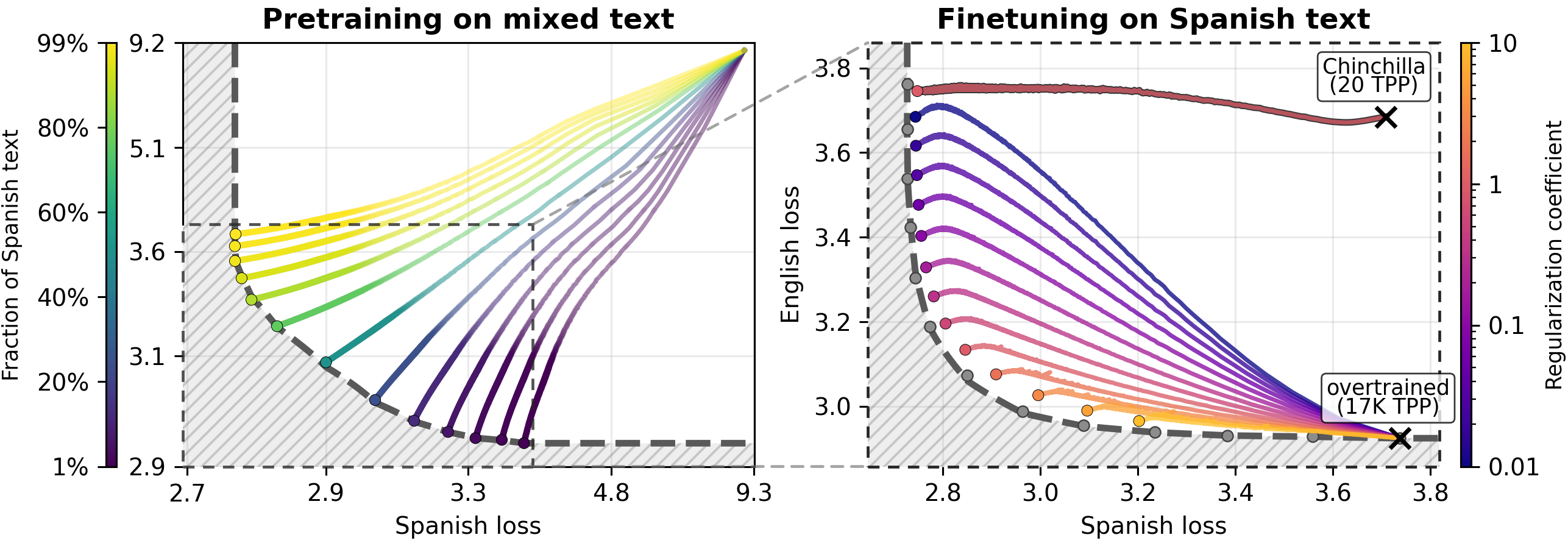}
    \caption{
    \textbf{Forgetting can be mitigated by regularizing on self-generated samples, but model capacity still lower-bounds forgetting.}
    \hspace{0mm}
    \textbf{Left:} We pretrain a small transformer language model on a mix of English and Spanish text until convergence. The color of each line indicates the \viridisEight{f}{r}{a}{c}{t}{i}{o}{n} of Spanish data. Because the model is small and trained to convergence, it does not have the capacity to achieve low loss on both languages simultaneously.
    \hspace{0mm}
    \textbf{Right:} We finetune two pretrained models (marked by crosses) on Spanish. The upper model is pretrained using 20 tokens per parameter (Chinchilla scaling), the lower on 17,000 (overtrained).
    To prevent forgetting during finetuning, we penalize the model's KL divergence on self-generated text.
    Because the Chinchilla-scaled model has plenty of spare capacity, it can improve its Spanish performance without degrading on English.
    In contrast, the overtrained model is near capacity, so it has to directly trade off learning and forgetting, which we control using the regularization \plasmaEleven{c}{o}{e}{f}{f}{i}{c}{i}{e}{n}{t}.
    }
    \label{fig:capacity}
\end{figure}

\section{Introduction}\label{sec:introduction}

Frontier language models are expected to do mathematical reasoning, use computer tools, and work with multimodal inputs, all in a single model. As user demands evolve, it is common practice to finetune these models to improve specific capabilities rather than to train models from scratch, especially as the cost of pretraining has become prohibitively expensive~\citep{gupta2023continualpretraininglargelanguage}. However, while finetuning improves performance on new data, it can also degrade the model's capabilities acquired during pretraining (commonly referred to as catastrophic forgetting~\citep{kirkpatrick2017overcoming}), which can extend beyond task-specific accuracy and affect the model’s robustness, reasoning, and even safety and alignment behavior~\citep{luo2025empiricalstudycatastrophicforgetting,bethune2025scaling,finetune_safety}. This tension creates a central challenge for foundational language models, which are expected to continually incorporate new capabilities without forgetting old ones~\citep{wang2024comprehensivesurveyforgettingdeep}. We thus ask a simple question: \emph{when models are trained on new data, when do they forget, and what is needed to prevent forgetting?}

\begin{figure}[H]
    \centering
    \includegraphics[width=\textwidth]{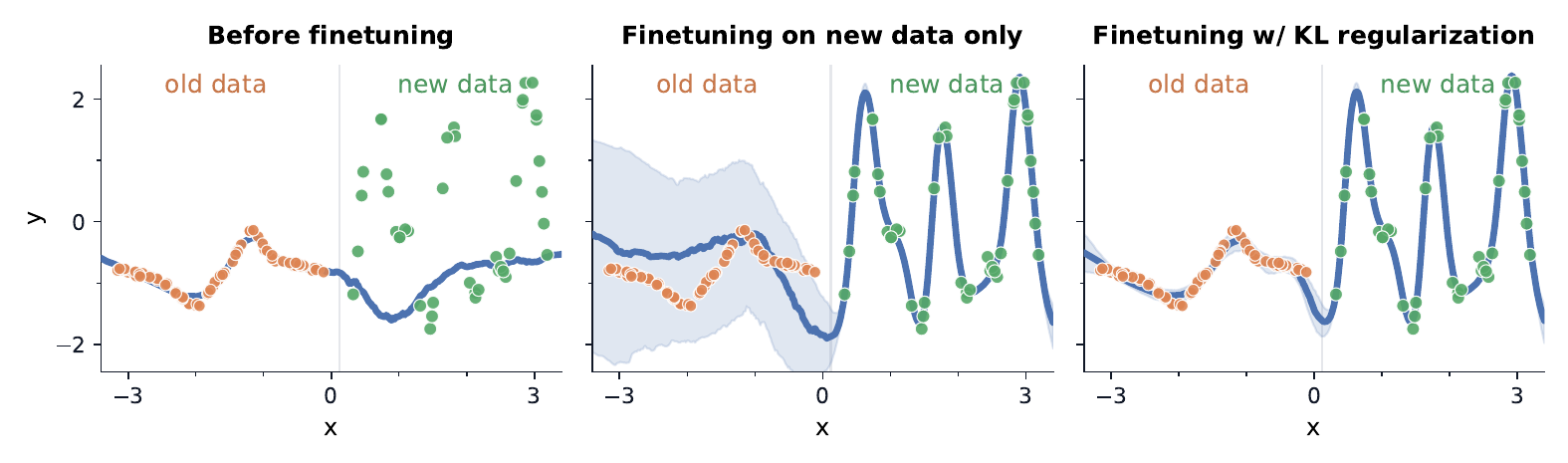}
    \caption{
    \textbf{Regularizing on past (replay) data prevents forgetting.}
    An MLP is first pretrained on data on the {\color{deep_orange!90!black}left} and then finetuned on data on the {\color{deep_green!85!black}right}.
    Without any regularization, training on new data changes the model's predictions on the old data, substantially degrading performance.
    Adding a KL divergence penalty on the prior data keeps the old predictions fixed while allowing the model to fit the new data. The shaded area represents a 50\% interquartile range across sampled models.
    }
    \label{fig:1d}
\end{figure}

We view forgetting as a change in the model's outputs on prior data~\citep{li2017learningforgetting,kotha2026replaying}. Under this view, a natural way to prevent forgetting is to regularize the model's predictions on prior samples to not change during finetuning, as we show in \Cref{fig:1d}. 
A practical limitation, however, is that pretraining data is often massive, proprietary, or unavailable. This limitation is less restrictive for language models, as they can directly generate samples that approximate the pretraining distribution, allowing new approaches that were previously challenging~\citep{masana2022class}.

The layout of this paper follows our key contributions:
\begin{itemize}[leftmargin=*, itemsep=0pt, parsep=0pt, topsep=0pt]
\item \Cref{sec:replay,sec:it}: We show that replay of \textbf{pretraining data greatly reduces forgetting} in both base and instruction-tuned language models.
If access to pretraining data is not available, it can be substituted without loss of performance by \textbf{self-generated} data from the model.
Additionally, constraining the \textbf{KL divergence} on replay data is more effective at preventing forgetting than using the standard next-token-prediction objective for regularization.
\item \Cref{sec:capacity}: While replay data is effective at reducing forgetting, we show there is a lower bound on forgetting, arising from \textbf{limited model capacity}. A model that is pretrained close to saturation (i.e., a small model pretrained on a large dataset) might not have sufficient capacity to absorb new information without overwriting prior information. We show that larger models trained on fewer tokens are easier to finetune, both with and without regularization.
\item \Cref{sec:lr}: Even when a model has sufficient capacity and retention signal (e.g., through regularization on pre-training data), forgetting still depends on how the model is optimized. We focus on learning rate because it directly controls both parameter drift and training cost. While low learning rates can reduce forgetting, they require many more optimizer steps. We show that replay data breaks this tradeoff, enabling \textbf{compute-efficient finetuning} with a high learning rate without significant forgetting.
\end{itemize}

Our code is available at  \url{https://github.com/martin-marek/forgetting}.

\section{Preventing Forgetting Using Replay Data}
\label{sec:replay}

Motivated by \Cref{fig:1d}, the primary method we use to reduce forgetting is to constrain the model's predictions on prior (``replay'' \citep{rolnick2019experiencereplaycontinuallearning})
data to not change during finetuning. This approach directly prevents forgetting: as long as the model outputs on prior data remain unchanged, then so should its performance on any tasks overlapping with prior data. This perspective challenges the commonly assumed tradeoff between learning and forgetting (also referred to as the stability-plasticity tradeoff \citep{mermillod2013stability}): a large model with sufficient capacity should be able to learn new data without changing its predictions on past data (hence without forgetting).

We implement the use of replay data by adding an auxiliary loss term $L_\mathrm{replay}$ to the training objective: $L_\mathrm{total} = L_\mathrm{downstream} + \lambda L_\mathrm{replay}$, where $\lambda$ dictates the regularization strength. In all of our experiments, the downstream loss is simply the next token prediction (NTP) loss, measured as the cross-entropy (negative log-likelihood) of downstream tokens. For the replay loss, we consider two different objectives. The first objective is the standard NTP loss on replay data. The second objective is the forward token-level Kullback-Leibler (KL) divergence, measured on replay data. When the replay data is generated by the base model, both of these objectives become equal in expectation to sequence-level forward KL divergence. Indeed, by denoting the base model $\pi$ and the downstream model $\theta$:
\begin{align}
\begin{split}
L_\mathrm{replay}^\mathrm{KL}
&= D_\mathrm{KL} \left( p_\pi (x) \| p_\theta (x) \right) \\
&= \mathbb{E}_{p_\pi(x)} \left[ \overset{\textcolor{gentlelilac}{\text{const. w.r.t. }\theta}}{\gentlelilaccancel{\log p_\pi(x)}} - \log p_\theta(x) \right] \\
&\overset{\mathclap{\textcolor{gentlelilac}{\text{const.}}}}{=} \; \; \mathbb{E}_{p_\pi(x)} \left[ - \log p_\theta(x) \right] \\
&= \mathbb{E}_{p_\pi(x)} \left[ L_\mathrm{replay}^\mathrm{NTP} \right] \ \textcolor{notegray}{\rightarrow\ \text{NTP under $p_\pi$ is a Monte Carlo estimator of KL}}
\label{eq:kl_vs_ntp}
\end{split}
\end{align}

\begin{figure}[H]
    \centering
    \includegraphics[width= \textwidth]{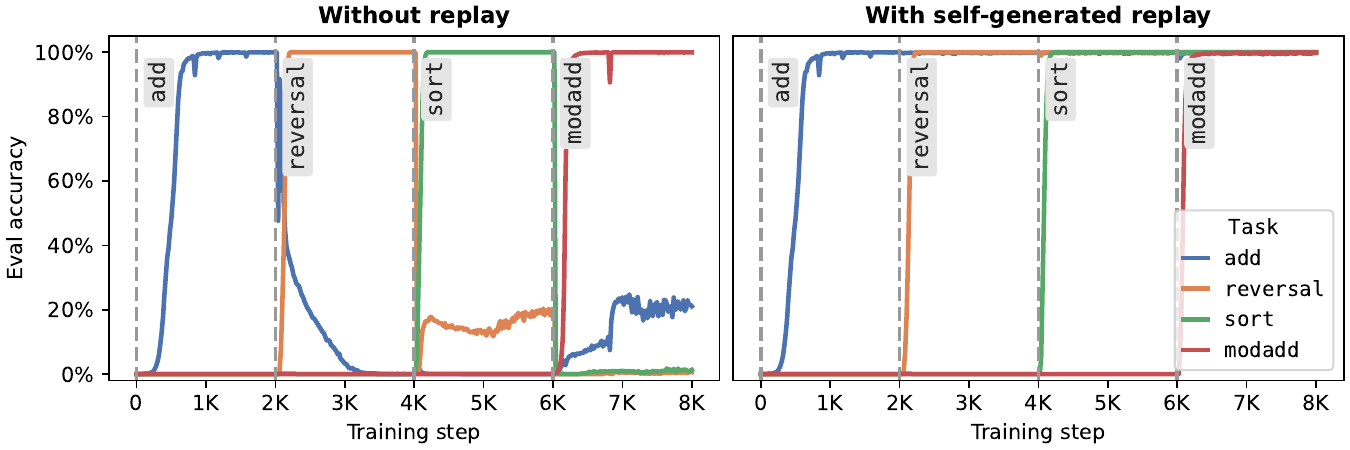}
    \caption{
    \textbf{Replay mitigates forgetting of language models under shifting tasks.} A small (2M) transformer language model is sequentially trained to perform \texttt{add}, \texttt{reversal}, \texttt{sort}, and \texttt{modadd} on 3-digit decimal inputs. \hspace{0mm} \textbf{Left:} Using standard training, as the model learns a new task, its accuracy on prior tasks completely degrades. \hspace{0mm} \textbf{Right:} By adding self-generated replay data, forgetting is entirely eliminated.
    }
    \label{fig_cl}
\end{figure}

We illustrate the effect of replay data using a toy continual learning experiment in \Cref{fig_cl}. A small language model with 2M parameters is sequentially trained on four simple tasks:
addition, digit reversal, digit sorting, and addition modulo $1000$.
The problems are written as short token sequences, for example,
addition looks like
\tok{add}\toksep\tok{3}\toksep\tok{4}\toksep\tok{7}\toksep\tok{|}\toksep
\tok{5}\toksep\tok{8}\toksep\tok{9}\toksep\tok{=}\toksep
\tok{0}\toksep\tok{9}\toksep\tok{3}\toksep\tok{6}.

For replay, we do not store any examples. Instead, before training on each task, we freeze a copy of the current model and sample examples of the earlier tasks from it. These generated examples look exactly like the training data for the old tasks, so training on this data preserves the model's capabilities on prior tasks while it learns the new task. 

These results illustrate that replay, even when using entirely self-generated data, can provide enough retention signal to completely preserve prior task performance during sequential training.

\begin{figure}[H]
    \centering
    \includegraphics[width=0.62\textwidth]{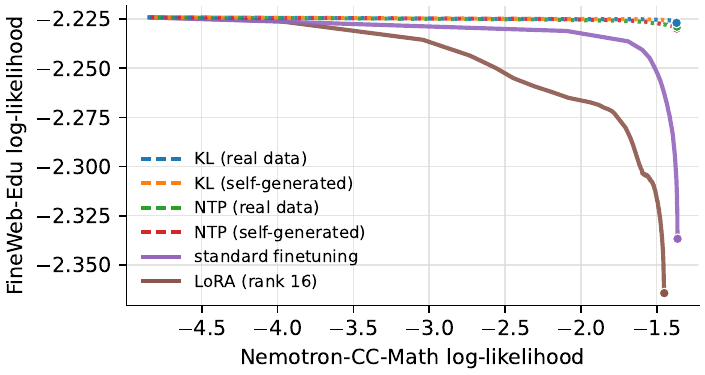}
    \caption{
    \textbf{Simple regularization mitigates forgetting in realistic continual learning (CL) scenarios.}
    A 205M parameter transformer language model is pretrained on 30B tokens of FineWeb-Edu and finetuned for multiple epochs on 10M tokens of Nemotron-CC-Math with early stopping. Each line shows a single training run. Both KL and NTP use a strong regularization coefficient ($\lambda=10$) to minimize forgetting. Across methods we use a fixed small learning rate $(10^{-5})$ to minimize forgetting; the effect of learning rate is further discussed in \Cref{sec:lr}.
    }
    \label{fig_method_comparison}
\end{figure}

\Cref{fig_method_comparison} tests replay in a language modeling setting. We first pretrain a transformer \citep{vaswani2017attention} on FineWeb-Edu (a general-domain pretraining corpus)~\citep{penedo2024finewebdatasetsdecantingweb} and then finetune it on Nemotron-CC-Math~\citep{mahabadi2025nemotronccmath133billiontokenscalehigh} to improve math performance. We compare standard finetuning, LoRA~\citep{lora}, and replay-based regularization.
Standard finetuning achieves strong downstream performance but forgets the pretraining distribution, while LoRA learns less and forgets more. KL regularization results in least forgetting (both with real and self-generated replay data), while NTP lags only slightly behind.

These results show that learning a new task does not necessitate forgetting, as long as the loss function includes past data. Conversely, just because LoRA learns less than full-finetuning does not mean that it forgets less. Rather than viewing learning and forgetting (``stability and plasticity'') as being inherently in tension~\citep{french1999catastrophic,mccloskey1989catastrophic}, we suggest that the tradeoff depends on whether the objective provides a retention signal for the prior distribution and whether the model has enough spare capacity to absorb new information (we discuss model capacity further in \Cref{sec:capacity}).

Importantly, our goal is not to propose a complex data-generation procedure or a sophisticated regularization method to prevent forgetting. Instead, we show that the model's own samples are sufficient to replace real pretraining data. Notably, unlike methods that require stored exemplars~\citep{kotha2026replaying} or prompt-conditioned generation~\citep{huang2024mitigatingcatastrophicforgettinglarge}, our synthetic replay data is generated directly from the model, without any prompting, making it a minimal form of replay: no stored memory buffers and no reliance on the model's in-context learning capabilities.

This finding raises a natural question: if regularization on replay data is so effective, when does forgetting still occur? We next show that replay is not the only factor we must consider. Rather, forgetting also depends on model capacity, dataset size, pretraining learning rate, and finetuning learning rate.

\section{Model Capacity is Necessary for Learning without Forgetting}
\label{sec:capacity}

In \Cref{sec:introduction,sec:replay}, regularizing on data drawn from the prior task distribution (both real and self-generated) almost entirely eliminated forgetting.
In this section, we show that learning without forgetting is only possible for models with sufficient capacity.

Any model has finite capacity to store information -- for modern transformer-decoder language models, the capacity is typically 2--4 bits per parameter \citep{physics_llms_capacity,llms_memorize}. While compute-optimal pretraining uses only 7--20 tokens per parameter \citep{chinchilla,scaling_second_order}, inference-optimized models are much smaller and trained for much longer, up to 60,000 tokens per parameter \citep{scaling_inference,qwen3}. Our central claim is that as models get smaller and are trained for longer, they approach their maximum capacity to store information during pretraining. Hence, in order for a capacity-constrained model to absorb new information during finetuning, it has to forget information learned during pretraining.

We illustrate the limited capacity of a small language model with 6M parameters by pretraining it on 100B tokens of text (17,000 tokens per parameter) in \Cref{fig:capacity}.
Because of the model's limited capacity, it is unable to achieve low loss on both pretraining and downstream data at the same time -- creating a direct tradeoff between learning and forgetting during finetuning.
In the rest of this section, we argue that this tradeoff is inherent to a limited model capacity. \textit{Models that are large or trained on few tokens are able to learn without significant forgetting}.

\begin{figure}[H]
    \centering
    \includegraphics[width=\textwidth]{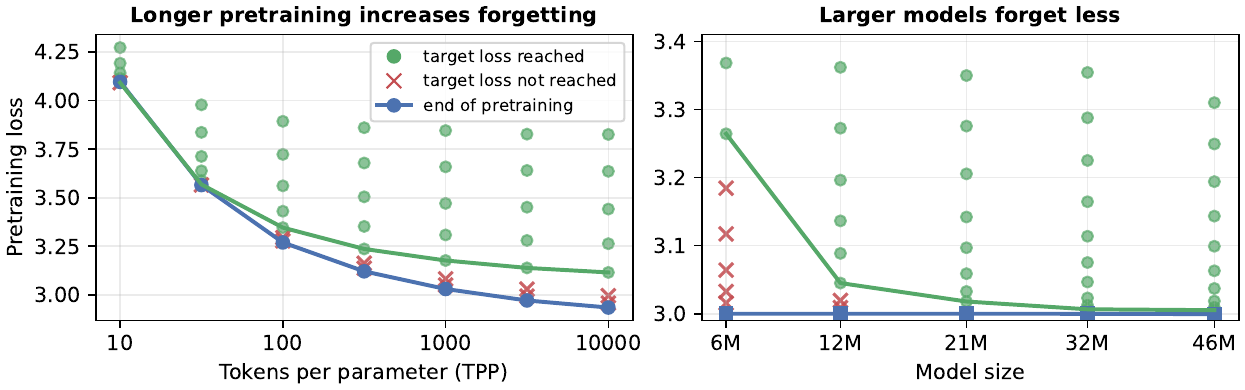}
    \caption{
    \textbf{Overtrained models forget more.}
    \hspace{0mm}
    We study the effect of model capacity by pretraining and finetuning on the English and Spanish subsets of C4 \citep{c4}. During finetuning, we sweep over different strengths of KL regularization -- each run is shown as a single point on the plot. We finetune each model either until it reaches a fixed target finetuning loss (\protect\greenpoint{}) or until the optimizer exceeds 100 epochs, at which point we consider the model failing to converge (\protect\redcross{}). Using weak regularization, every pretrained model is able to achieve the target finetuning loss, although with severe forgetting. Conversely, strong KL regularization reduces forgetting, but small / overtrained models fail to reach the target finetuning loss.
    \hspace{0mm}
    \textbf{Left:} Given a fixed-size model (6M parameters), longer pretraining reduces spare capacity in the model after pretraining, resulting in increased forgetting.
    \hspace{0mm}
    \textbf{Right:}~Given models of different sizes (all pretrained until the same pretraining loss), larger models forget less, since they have more spare capacity available.
    \hspace{0mm}
    }
    \label{fig:capacity_detailed}
\end{figure}

\begin{figure}[H]
    \centering
    \vspace{-4mm}
    \includegraphics[width=0.8 \textwidth]{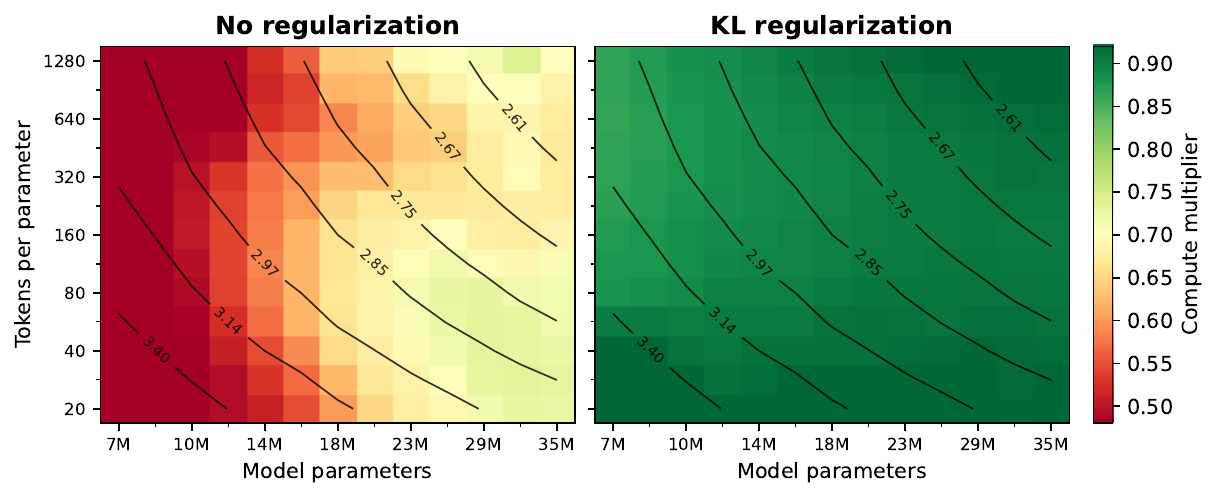}
    \caption{
    \textbf{Model capacity affects forgetting, but replay is more important.}
    \hspace{0mm}
    We pretrain models of varying sizes on a varying number of tokens of FineWeb-Edu, then finetune them on Nemotron-CC-Math until a fixed target loss is reached.
    The contour lines show pretraining loss before finetuning. The colorbar expresses pretraining loss after finetuning as a compute multiplier (a compute multiplier of 90\% means that pretraining loss after finetuning is the same as if the model was pretrained on 10\% fewer tokens).
    \hspace{0mm}
    \textbf{Left:} without any regularization, given a fixed pretraining loss (contour line), larger models forget less. 
	\hspace{0mm}
    \textbf{Right:} adding KL regularization on pretraining data (with a fixed coefficient $\lambda=10$) mostly eliminates forgetting, to a much larger extent than model size alone. Still, for a given pretraining loss, larger models forget less.
    }
    \label{fig_modelsize_tpp}
\end{figure}

To test the effect of model and dataset size more directly, we consider two complementary settings in \Cref{fig:capacity_detailed}: pretraining a fixed-size model for a varying number of tokens (left), and pretraining models of varying sizes to achieve the same pretraining loss (right). We again see that models pretrained close to saturation are unable to absorb new information without forgetting. \Cref{fig_modelsize_tpp} shows that model size affects forgetting both with and without replay, although the effect of replay is more significant than the effect of model size.

\section{Learning Rate: Training Time and Forgetting}
\label{sec:lr}

\begin{figure}[H]
    \centering
    \includegraphics[width=\textwidth]{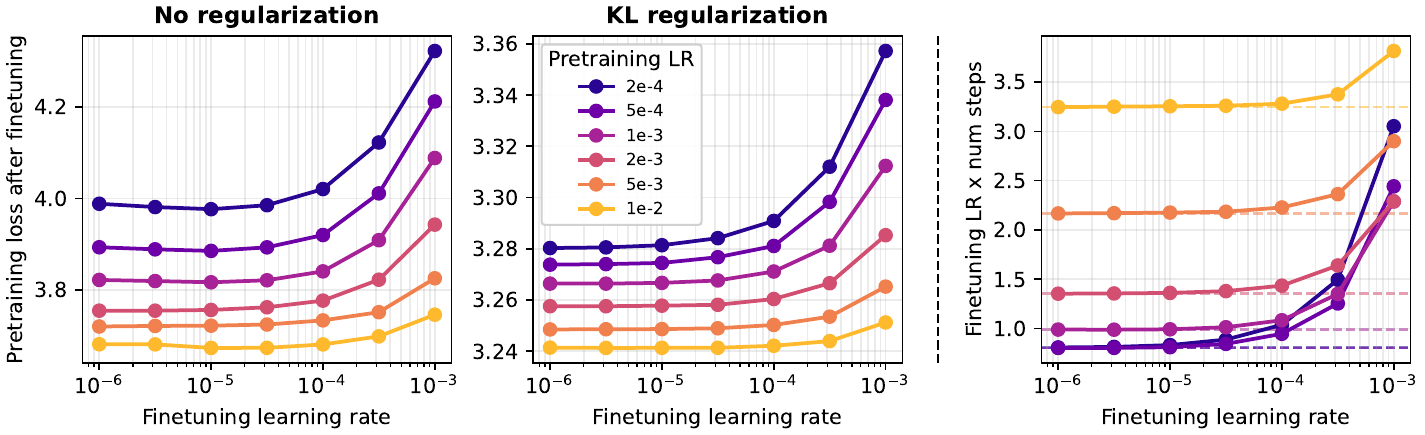}
    \caption{
    \textbf{Pretraining and finetuning learning rate affect forgetting.}
    We pretrain small (6M parameters) language models using different learning rates until they reach the target loss of 3.2 nats on English text, then finetune them using different learning rates until they reach the same loss on Spanish text.
    \hspace{0mm}
    \textbf{Left / Middle:}
    Both with and without replay data (KL regularization), using a high pretraining learning rate and a low finetuning learning rate reduces forgetting. 
    \hspace{0mm}
    \textbf{Right:}
    There are diminishing returns to decreasing the finetuning learning rate. As the learning rate decreases, the training dynamics converge to a flow process \citep{adam_flow}, leading to an inversely proportional number of training steps to achieve a fixed downstream loss (learning rate times number of steps is constant), without any benefits in model performance.
    }
    \label{fig:learning_rate}
\end{figure}

We now study forgetting through the lens of optimization. Previous works have already shown that a low learning rate in finetuning helps preserve performance on prior tasks \citep{loss_curvature}, and a high learning rate in pretraining leads to flatter minima \citep{central_flow}, which can lead to improved downstream performance \citep{yano2026pretrainingllmlearningrate,lr_decay_quantization}. We verify these results in \Cref{fig:learning_rate} and extend them to the data replay setting.

\begin{figure}[H]
    \centering
    \includegraphics[width=0.6 \textwidth]{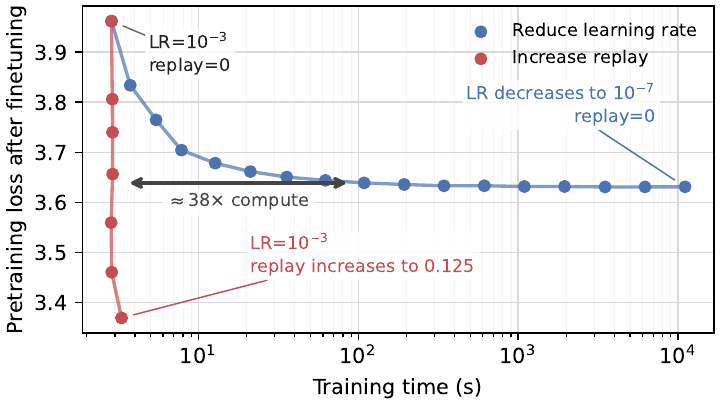}
    \caption{\textbf{Replay enables compute-efficient high-learning-rate finetuning.} We study a compute-efficient approach to finetuning on the English and Spanish subsets of the C4 dataset. Replay allows the model to be finetuned with a high learning rate while minimizing forgetting, reducing the number of optimization steps to reach the downstream target loss, thereby reducing wall time.
    }
    \label{fig:high_lr}
\end{figure}

These results suggest a practical tradeoff: smaller finetuning learning rates reduce forgetting, but require more optimization steps to reach the same downstream target loss, inversely increasing compute. \Cref{fig:high_lr} shows that replay data breaks this tradeoff. We use the same setup as \Cref{fig:learning_rate}, but report wall-clock time required to reach the downstream target loss.
To minimize wall time, we implement replay by mixing downstream and replay sequences inside a single batch and compute a single NTP loss value on this mixed batch.\footnote{All other experiments use one batch of downstream-only data and one batch of replay-only data.}
With replay, the model can use a high finetuning learning rate to reach the target loss much faster, while avoiding the forgetting that high learning rates otherwise induce.

\section{Instruction-Tuned Models}
\label{sec:it}

While \Cref{sec:introduction,sec:replay,sec:capacity,sec:lr} studied base language models pretrained from scratch, these results do not immediately transfer to the language models that are used in practice. One difference is that pretraining data for most open-weight LLMs is not public, forcing us to rely on self-generated  replay data. An even larger difference is that most practical applications rely on instruction-tuned (IT) language models that follow a user-assistant chat template. This poses a challenge: how should we generate samples from an instruction-tuned model? IT models are typically trained with loss masking, so they may not be able to generate user prompts; they can only reliably generate an assistant response given a user prompt. However, if we wanted to generate replay data by repeatedly prompting the model, it is not clear how we could collect a representative sample of prompts that covers the whole pretraining distribution of the model.

Rather than adopting previous practices (e.g., stored examples, task descriptions, or handcrafted prompts), we approach this problem by prompting the IT model only with a single \tok{BOS} token, imitating the format of the model's \textit{pretraining} (rather than post-training) data. This approach works surprisingly well for making Llama-3.2-1B-Instruct \citep{llama3} generate pretraining-like data, even \textit{after} the model has been instruction tuned. For example, here are the first three sequences sampled from Llama-3.2-1B-Instruct prompted with \tok{BOS}:
\begin{itemize}
    \item \mbox{{\fontsize{9.6}{9}\selectfont\texttt{[BOS]Title:\ A randomized controlled trial of a new, evidence-based...}}}
    \item \texttt{[BOS]The classic tale of Cinderella has its roots in a medieval...}
    \item \texttt{[BOS]import numpy as np \ensuremath{\hookleftarrow} from scipy.optimize import minimize...}
\end{itemize}

We study the setting of finetuning Llama-3.2-1B-Instruct on Verilog -- a hardware description language that is likely under-represented in the model's training data \citep{pinckney2025revisiting}. This task is challenging for the model, because it has to learn a new modeling language. We test using replay data self-generated by the model (by prompting it with \tok{BOS}), as well as substitute pretraining data from OLMo3 \citep{OLMo3}, since the original Llama 3 pretraining data is not publicly available. We finetune Llama to generate Verilog code given text descriptions of the code as the prompt. We use a cleaned subset of PyraNet-Verilog \citep{verilog}, consisting of around 200K examples. We measure performance on Verilog using next-token prediction accuracy on a held-out sample of the dataset. We measure forgetting as the accuracy of the model averaged across three science and general knowledge benchmarks: MMLU \citep{mmlu}, CommonsenseQA \citep{commonsenseqa}, and ARC-Challenge \citep{arc}.

\begin{figure}[H]
    \centering
    \includegraphics[width=0.65 \textwidth]{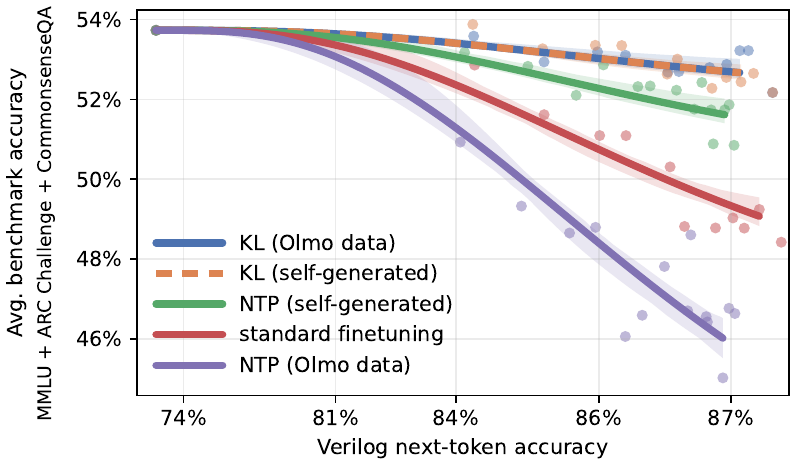}
    \caption{
    \textbf{Instruction-tuned models benefit from replay of pretraining data.}
    We finetune Llama-3.2-1B-Instruct to generate Verilog code.
    Standard finetuning improves downstream performance at the cost of forgetting, while KL regularization almost entirely eliminates forgetting. KL regularization works equally well with both substitute and self-generated data, whereas NTP regularization is sensitive to the distribution of replay data.
    }
    \label{fig:instruct}
\end{figure}

\Cref{fig:instruct} shows that penalizing KL divergence on both substitute and self-generated data almost completely eliminates forgetting. In contrast, the standard next-token-prediction loss on substitute data actually increases forgetting compared to standard finetuning. A possible explanation for this result is that NTP loss on OLMo data forces the model to learn data from an entirely new distribution, thereby exacerbating forgetting. In contrast, a KL divergence penalty merely enforces that the predictions of the model do not change.

It is also worth noting that on self-generated replay data, the NTP and KL objectives have the same gradient in expectation, as shown in \Cref{eq:kl_vs_ntp}. But while the expectation is the same, NTP on self-generated data yields more noisy gradients, treating each sampled token as a hard target. KL instead uses the full next-token distribution of the reference model, replacing a single-token target with a sum over the vocabulary.\footnote{This is the same difference as soft vs.\ hard knowledge distillation.} Therefore, we can attribute the different performance between NTP and KL regularization to a small batch size and the higher gradient variance of NTP compared to KL.

\section{Related Work}
\label{sec:related_works}

\paragraph{Continual Learning.}

Continual learning (CL) studies how models can adapt to new data streams without overwriting previously acquired knowledge. This problem is commonly described as catastrophic forgetting~\citep{kirkpatrick2017overcoming} and is closely tied to the stability-plasticity dilemma: a model must remain plastic enough to acquire new information while remaining stable enough to preserve past knowledge~\citep{mermillod2013stability}. Classical CL methods address this tension through parameter-space regularization~\citep{kirkpatrick2017overcoming,aljundi2018memory}, rehearsal of stored replay data~\citep{rolnick2019experiencereplaycontinuallearning,rebuffi2017icarlincrementalclassifierrepresentation,shin2017continuallearningdeepgenerative}, and architectural expansion~\citep{adila2026grow,der,foster}, largely motivated by settings where access to past data is limited or costly. Recent work highlights a key trade-off in continual learning: while large replay budgets effectively prevent catastrophic forgetting (stability)~\citep{prabhu2023computationally}, excessive replay limits the model's capacity to learn new tasks (plasticity)~\citep{cho2026forgetforgettingcontinuallearning,dohare2024loss}. Large language models inherit these challenges but also change the setting: forgetting may affect broad capabilities rather than only task accuracy, task boundaries are often weakly defined, and replay need not consist of exact stored examples~\citep{mitchell2025continual,wu2024continuallearninglargelanguage}. While prior work has studied individual mechanisms for mitigating forgetting, such as replay or regularization, we ask when these mechanisms are sufficient and what factors determine their effectiveness, focusing on the interaction between retained data, model capacity, and optimization.
 
\paragraph{Self-generated Replay.}

The idea of using self-generated samples for continual learning is not novel~\citep{shin2017continuallearningdeepgenerative,cho2025peer,hcl}, and recent work has applied related replay methods to large language models~\citep{huang2024mitigatingcatastrophicforgettinglarge,resta2024selfgeneratedreplaymemoriescontinual}. These approaches often rely on stored examples, prompts, or the model's in-context learning ability. In contrast, we generate replay directly from the frozen reference model without storing past data, making the method applicable when access to pretraining data is unavailable and even when the model is too weak for reliable in-context generation.

\paragraph{Mixing Pretraining and Finetuning Data.}

A closely related line of work studies how pretraining and finetuning data can be combined both during pretraining and during finetuning \citep{ke2023continualpretraininglanguagemodels}. \citet{bethune2025scaling} derive scaling laws for forgetting under pretraining-data injection, while \citet{kotha2026replaying} show that replaying pretraining data during finetuning can improve target-domain data efficiency rather than merely preserving the general performance of the base model. A complementary direction introduces finetuning data earlier in pretraining to improve later  finetuning~\citep{baek2026finetunersfallacypretrainfinetuning,korbak2023pretraininglanguagemodelshuman}. In contrast, we ask how data from the pretraining distribution should be used during finetuning to preserve the model's prior capabilities, and show that substitute or self-generated samples can be effective when the original pretraining data is unavailable.

\paragraph{Effect of Capacity and Optimization in Continual Learning.}

Several works suggest that pretraining conditions can affect finetuning behavior. \citet{springer2025overtrainedlanguagemodelsharder} report that models pretrained for longer can become harder to finetune, and multiple works demonstrate that models with similar pretraining loss can differ in downstream performance due to optimization bias during pretraining \citep{liu2022pretraininglossbetterdownstream,sharpness_aware_pretraining,yano2026pretrainingllmlearningrate,lr_decay_quantization}. Other works connect these behaviors to model capacity: \citet{kim2024knowledge} show that model capacity affects learning and forgetting. In a complementary direction, we extend this perspective to finetuning with replay, showing that regularization can greatly reduce forgetting in most settings, but fails once the model has too little spare capacity. Learning rate provides another perspective: \citet{yano2026pretrainingllmlearningrate} show that pretraining without learning-rate decay can improve finetuning performance, while \citet{lr_decay_quantization} illustrate that training dynamics can also affect post-training quantization. Together, these works study different aspects of how pretraining and optimization affect later adaptation. Our work brings these factors together, showing how model size, pretraining length, learning rate, and replay regularization jointly determine whether a model can continually learn without forgetting.

\section{Conclusion}
We study when language models forget during finetuning and what is needed to prevent it. Our results suggest that forgetting is not an unavoidable consequence of learning new data, but a consequence of drift on the pretraining distribution. Regularizing on pretraining data greatly reduces forgetting, and in language models, this data can be effectively self-generated, even in an instruction-tuned model. However, learning without forgetting requires sufficient model capacity. This perspective turns forgetting into a controllable outcome of data, model capacity, and optimization choices, and enables compute-efficient high-learning-rate finetuning that preserves prior capabilities.

\paragraph{Limitations.}
Most experiments presented in this paper involve training on a single new task; multi-task continual learning could pose new challenges, such as loss of plasticity \citep{dohare2024loss}. To be able to overtrain models, run exhaustive sweeps over hyperparameters, and use very small learning rates, we worked with small models (up to 46M parameters) for the majority of the experiments; only \Cref{fig:instruct} uses a model at the billion-parameter scale. Lastly, our experiments rely on proxies for measuring model capacity, such as model size and pretraining loss, rather than directly measuring the information stored in the model.

\section*{Acknowledgements}
We thank Alexandra Souly for helpful discussions. This research was supported by NSF CAREER IIS-2145492, DARPA AIQ HR00112590066, and Google’s TPU Research Cloud (TRC) program: \url{https://sites.research.google/trc/}.

\bibliographystyle{unsrtnat}
\bibliography{main}

\begin{thebibliography}{64}
\providecommand{\natexlab}[1]{#1}
\providecommand{\url}[1]{\texttt{#1}}
\expandafter\ifx\csname urlstyle\endcsname\relax
  \providecommand{\doi}[1]{doi: #1}\else
  \providecommand{\doi}{doi: \begingroup \urlstyle{rm}\Url}\fi

\bibitem[Gupta et~al.(2023)Gupta, Thérien, Ibrahim, Richter, Anthony, Belilovsky, Rish, and Lesort]{gupta2023continualpretraininglargelanguage}
Kshitij Gupta, Benjamin Thérien, Adam Ibrahim, Mats~L. Richter, Quentin Anthony, Eugene Belilovsky, Irina Rish, and Timothée Lesort.
\newblock Continual pre-training of large language models: How to (re)warm your model?, 2023.
\newblock URL \url{https://arxiv.org/abs/2308.04014}.

\bibitem[Kirkpatrick et~al.(2017)Kirkpatrick, Pascanu, Rabinowitz, Veness, Desjardins, Rusu, Milan, Quan, Ramalho, Grabska-Barwinska, Hassabis, Clopath, Kumaran, and Hadsell]{kirkpatrick2017overcoming}
James Kirkpatrick, Razvan Pascanu, Neil Rabinowitz, Joel Veness, Guillaume Desjardins, Andrei~A. Rusu, Kieran Milan, John Quan, Tiago Ramalho, Agnieszka Grabska-Barwinska, Demis Hassabis, Claudia Clopath, Dharshan Kumaran, and Raia Hadsell.
\newblock Overcoming catastrophic forgetting in neural networks.
\newblock \emph{Proceedings of the National Academy of Sciences}, 114\penalty0 (13):\penalty0 3521–3526, March 2017.
\newblock ISSN 1091-6490.
\newblock \doi{10.1073/pnas.1611835114}.
\newblock URL \url{http://dx.doi.org/10.1073/pnas.1611835114}.

\bibitem[Luo et~al.(2025)Luo, Yang, Meng, Li, Zhou, and Zhang]{luo2025empiricalstudycatastrophicforgetting}
Yun Luo, Zhen Yang, Fandong Meng, Yafu Li, Jie Zhou, and Yue Zhang.
\newblock An empirical study of catastrophic forgetting in large language models during continual fine-tuning, 2025.
\newblock URL \url{https://arxiv.org/abs/2308.08747}.

\bibitem[Bethune et~al.(2025)Bethune, Grangier, Busbridge, Gualdoni, Cuturi, and Ablin]{bethune2025scaling}
Louis Bethune, David Grangier, Dan Busbridge, Eleonora Gualdoni, Marco Cuturi, and Pierre Ablin.
\newblock Scaling laws for forgetting during finetuning with pretraining data injection, 2025.
\newblock URL \url{https://arxiv.org/abs/2502.06042}.

\bibitem[Qi et~al.(2024)Qi, Zeng, Xie, Chen, Jia, Mittal, and Henderson]{finetune_safety}
Xiangyu Qi, Yi~Zeng, Tinghao Xie, Pin-Yu Chen, Ruoxi Jia, Prateek Mittal, and Peter Henderson.
\newblock Fine-tuning aligned language models compromises safety, even when users do not intend to!
\newblock In \emph{International Conference on Learning Representations}, volume 2024, pages 30988--31043, 2024.

\bibitem[Wang et~al.(2024)Wang, Yang, Shen, and Huang]{wang2024comprehensivesurveyforgettingdeep}
Zhenyi Wang, Enneng Yang, Li~Shen, and Heng Huang.
\newblock A comprehensive survey of forgetting in deep learning beyond continual learning, 2024.
\newblock URL \url{https://arxiv.org/abs/2307.09218}.

\bibitem[Li and Hoiem(2017)]{li2017learningforgetting}
Zhizhong Li and Derek Hoiem.
\newblock Learning without forgetting, 2017.
\newblock URL \url{https://arxiv.org/abs/1606.09282}.

\bibitem[Kotha and Liang(2026)]{kotha2026replaying}
Suhas Kotha and Percy Liang.
\newblock Replaying pre-training data improves fine-tuning, 2026.
\newblock URL \url{https://arxiv.org/abs/2603.04964}.

\bibitem[Masana et~al.(2022)Masana, Liu, Twardowski, Menta, Bagdanov, and van~de Weijer]{masana2022class}
Marc Masana, Xialei Liu, Bartlomiej Twardowski, Mikel Menta, Andrew~D. Bagdanov, and Joost van~de Weijer.
\newblock Class-incremental learning: survey and performance evaluation on image classification, 2022.
\newblock URL \url{https://arxiv.org/abs/2010.15277}.

\bibitem[Rolnick et~al.(2019)Rolnick, Ahuja, Schwarz, Lillicrap, and Wayne]{rolnick2019experiencereplaycontinuallearning}
David Rolnick, Arun Ahuja, Jonathan Schwarz, Timothy~P. Lillicrap, and Greg Wayne.
\newblock Experience replay for continual learning, 2019.
\newblock URL \url{https://arxiv.org/abs/1811.11682}.

\bibitem[Mermillod et~al.(2013)Mermillod, Bugaiska, and BONIN]{mermillod2013stability}
Martial Mermillod, Aurélia Bugaiska, and Patrick BONIN.
\newblock The stability-plasticity dilemma: investigating the continuum from catastrophic forgetting to age-limited learning effects.
\newblock \emph{Frontiers in Psychology}, Volume 4 - 2013, 2013.
\newblock ISSN 1664-1078.
\newblock \doi{10.3389/fpsyg.2013.00504}.
\newblock URL \url{https://www.frontiersin.org/journals/psychology/articles/10.3389/fpsyg.2013.00504}.

\bibitem[Vaswani et~al.(2017)Vaswani, Shazeer, Parmar, Uszkoreit, Jones, Gomez, Kaiser, and Polosukhin]{vaswani2017attention}
Ashish Vaswani, Noam Shazeer, Niki Parmar, Jakob Uszkoreit, Llion Jones, Aidan~N Gomez, {\L}ukasz Kaiser, and Illia Polosukhin.
\newblock Attention is all you need.
\newblock \emph{Advances in neural information processing systems}, 30, 2017.

\bibitem[Penedo et~al.(2024)Penedo, Kydlíček, allal, Lozhkov, Mitchell, Raffel, Werra, and Wolf]{penedo2024finewebdatasetsdecantingweb}
Guilherme Penedo, Hynek Kydlíček, Loubna~Ben allal, Anton Lozhkov, Margaret Mitchell, Colin Raffel, Leandro~Von Werra, and Thomas Wolf.
\newblock The fineweb datasets: Decanting the web for the finest text data at scale, 2024.
\newblock URL \url{https://arxiv.org/abs/2406.17557}.

\bibitem[Mahabadi et~al.(2025)Mahabadi, Satheesh, Prabhumoye, Patwary, Shoeybi, and Catanzaro]{mahabadi2025nemotronccmath133billiontokenscalehigh}
Rabeeh~Karimi Mahabadi, Sanjeev Satheesh, Shrimai Prabhumoye, Mostofa Patwary, Mohammad Shoeybi, and Bryan Catanzaro.
\newblock Nemotron-cc-math: A 133 billion-token-scale high quality math pretraining dataset, 2025.
\newblock URL \url{https://arxiv.org/abs/2508.15096}.

\bibitem[Hu et~al.(2021)Hu, Shen, Wallis, Allen-Zhu, Li, Wang, Wang, and Chen]{lora}
Edward~J. Hu, Yelong Shen, Phillip Wallis, Zeyuan Allen-Zhu, Yuanzhi Li, Shean Wang, Lu~Wang, and Weizhu Chen.
\newblock Lora: Low-rank adaptation of large language models, 2021.
\newblock URL \url{https://arxiv.org/abs/2106.09685}.

\bibitem[French(1999)]{french1999catastrophic}
Robert~M French.
\newblock Catastrophic forgetting in connectionist networks.
\newblock \emph{Trends in cognitive sciences}, 3\penalty0 (4):\penalty0 128--135, 1999.

\bibitem[McCloskey and Cohen(1989)]{mccloskey1989catastrophic}
Michael McCloskey and Neal~J Cohen.
\newblock Catastrophic interference in connectionist networks: The sequential learning problem.
\newblock In \emph{Psychology of learning and motivation}, volume~24, pages 109--165. Elsevier, 1989.

\bibitem[Huang et~al.(2024)Huang, Cui, Wang, Yang, Liao, Song, Yao, and Su]{huang2024mitigatingcatastrophicforgettinglarge}
Jianheng Huang, Leyang Cui, Ante Wang, Chengyi Yang, Xinting Liao, Linfeng Song, Junfeng Yao, and Jinsong Su.
\newblock Mitigating catastrophic forgetting in large language models with self-synthesized rehearsal, 2024.
\newblock URL \url{https://arxiv.org/abs/2403.01244}.

\bibitem[Allen-Zhu and Li(2024)]{physics_llms_capacity}
Zeyuan Allen-Zhu and Yuanzhi Li.
\newblock Physics of language models: Part 3.3, knowledge capacity scaling laws, 2024.
\newblock URL \url{https://arxiv.org/abs/2404.05405}.

\bibitem[Morris et~al.(2025)Morris, Sitawarin, Guo, Kokhlikyan, Suh, Rush, Chaudhuri, and Mahloujifar]{llms_memorize}
John~X. Morris, Chawin Sitawarin, Chuan Guo, Narine Kokhlikyan, G.~Edward Suh, Alexander~M. Rush, Kamalika Chaudhuri, and Saeed Mahloujifar.
\newblock How much do language models memorize?, 2025.
\newblock URL \url{https://arxiv.org/abs/2505.24832}.

\bibitem[Hoffmann et~al.(2022)Hoffmann, Borgeaud, Mensch, Buchatskaya, Cai, Rutherford, Casas, Hendricks, Welbl, Clark, et~al.]{chinchilla}
Jordan Hoffmann, Sebastian Borgeaud, Arthur Mensch, Elena Buchatskaya, Trevor Cai, Eliza Rutherford, DDL Casas, Lisa~Anne Hendricks, Johannes Welbl, Aidan Clark, et~al.
\newblock Training compute-optimal large language models.
\newblock \emph{arXiv preprint arXiv:2203.15556}, 10, 2022.
\newblock URL \url{https://arxiv.org/abs/2203.15556}.

\bibitem[Qiu et~al.(2026)Qiu, Chen, Phan, Lei, and Wilson]{scaling_second_order}
Shikai Qiu, Zixi Chen, Hoang Phan, Qi~Lei, and Andrew~Gordon Wilson.
\newblock Hyperparameter transfer enables consistent gains of matrix-preconditioned optimizers across scales.
\newblock In \emph{The Thirty-ninth Annual Conference on Neural Information Processing Systems}, 2026.
\newblock URL \url{https://openreview.net/forum?id=Ei6IsmxYrb}.

\bibitem[H{\"a}gele et~al.(2024)H{\"a}gele, Bakouch, Kosson, allal, Werra, and Jaggi]{scaling_inference}
Alexander H{\"a}gele, Elie Bakouch, Atli Kosson, Loubna~Ben allal, Leandro~Von Werra, and Martin Jaggi.
\newblock Scaling laws and compute-optimal training beyond fixed training durations.
\newblock In \emph{The Thirty-eighth Annual Conference on Neural Information Processing Systems}, 2024.
\newblock URL \url{https://openreview.net/forum?id=Y13gSfTjGr}.

\bibitem[Yang et~al.(2025)Yang, Li, Yang, Zhang, Hui, Zheng, Yu, Gao, Huang, Lv, et~al.]{qwen3}
An~Yang, Anfeng Li, Baosong Yang, Beichen Zhang, Binyuan Hui, Bo~Zheng, Bowen Yu, Chang Gao, Chengen Huang, Chenxu Lv, et~al.
\newblock Qwen3 technical report.
\newblock \emph{arXiv preprint arXiv:2505.09388}, 2025.
\newblock URL \url{https://arxiv.org/abs/2505.09388}.

\bibitem[Raffel et~al.(2023)Raffel, Shazeer, Roberts, Lee, Narang, Matena, Zhou, Li, and Liu]{c4}
Colin Raffel, Noam Shazeer, Adam Roberts, Katherine Lee, Sharan Narang, Michael Matena, Yanqi Zhou, Wei Li, and Peter~J. Liu.
\newblock Exploring the limits of transfer learning with a unified text-to-text transformer, 2023.
\newblock URL \url{https://arxiv.org/abs/1910.10683}.

\bibitem[Kalra et~al.(2026)Kalra, Gagnon-Audet, Gromov, Mediratta, Niu, Miller, and Shvartsman]{loss_curvature}
Dayal~Singh Kalra, Jean-Christophe Gagnon-Audet, Andrey Gromov, Ishita Mediratta, Kelvin Niu, Alexander~H Miller, and Michael Shvartsman.
\newblock A scalable measure of loss landscape curvature for analyzing the training dynamics of llms, 2026.
\newblock URL \url{https://arxiv.org/abs/2601.16979}.

\bibitem[Cohen et~al.(2025)Cohen, Damian, Talwalkar, Kolter, and Lee]{central_flow}
Jeremy Cohen, Alex Damian, Ameet Talwalkar, J~Zico Kolter, and Jason~D. Lee.
\newblock Understanding optimization in deep learning with central flows.
\newblock In \emph{The Thirteenth International Conference on Learning Representations}, 2025.
\newblock URL \url{https://openreview.net/forum?id=sIE2rI3ZPs}.

\bibitem[Yano et~al.(2026)Yano, Kiyono, Kobayashi, Takase, and Suzuki]{yano2026pretrainingllmlearningrate}
Kazuki Yano, Shun Kiyono, Sosuke Kobayashi, Sho Takase, and Jun Suzuki.
\newblock Pre-training llm without learning rate decay enhances supervised fine-tuning, 2026.
\newblock URL \url{https://arxiv.org/abs/2603.16127}.

\bibitem[Catalan-Tatjer et~al.(2026)Catalan-Tatjer, Ajroldi, and Geiping]{lr_decay_quantization}
Albert Catalan-Tatjer, Niccolò Ajroldi, and Jonas Geiping.
\newblock Training dynamics impact post-training quantization robustness, 2026.
\newblock URL \url{https://arxiv.org/abs/2510.06213}.

\bibitem[Ma et~al.(2021)Ma, Wu, and E]{adam_flow}
Chao Ma, Lei Wu, and Weinan E.
\newblock A qualitative study of the dynamic behavior for adaptive gradient algorithms, 2021.
\newblock URL \url{https://arxiv.org/abs/2009.06125}.

\bibitem[Grattafiori et~al.(2024)Grattafiori, Dubey, Jauhri, Pandey, Kadian, Al-Dahle, Letman, Mathur, Schelten, Vaughan, et~al.]{llama3}
Aaron Grattafiori, Abhimanyu Dubey, Abhinav Jauhri, Abhinav Pandey, Abhishek Kadian, Ahmad Al-Dahle, Aiesha Letman, Akhil Mathur, Alan Schelten, Alex Vaughan, et~al.
\newblock The llama 3 herd of models.
\newblock \emph{arXiv preprint arXiv:2407.21783}, 2024.
\newblock URL \url{https://arxiv.org/abs/2407.21783}.

\bibitem[Pinckney et~al.(2025)Pinckney, Batten, Liu, Ren, and Khailany]{pinckney2025revisiting}
Nathaniel Pinckney, Christopher Batten, Mingjie Liu, Haoxing Ren, and Brucek Khailany.
\newblock Revisiting verilogeval: A year of improvements in large-language models for hardware code generation, 2025.
\newblock URL \url{https://arxiv.org/abs/2408.11053}.

\bibitem[Olmo et~al.(2026)Olmo, :, Ettinger, Bertsch, Kuehl, Graham, Heineman, Groeneveld, Brahman, Timbers, Ivison, Morrison, Poznanski, Lo, Soldaini, Jordan, Chen, Noukhovitch, Lambert, Walsh, Dasigi, Berry, Malik, Shah, Geng, Arora, Gupta, Anderson, Xiao, Murray, Romero, Graf, Asai, Bhagia, Wettig, Liu, Rangapur, Anastasiades, Huang, Schwenk, Trivedi, Magnusson, Lochner, Liu, Miranda, Sap, Morgan, Schmitz, Guerquin, Wilson, Huff, Bras, Xin, Shao, Skjonsberg, Shen, Li, Wilde, Pyatkin, Merrill, Chang, Gu, Zeng, Sabharwal, Zettlemoyer, Koh, Farhadi, Smith, and Hajishirzi]{OLMo3}
Team Olmo, :, Allyson Ettinger, Amanda Bertsch, Bailey Kuehl, David Graham, David Heineman, Dirk Groeneveld, Faeze Brahman, Finbarr Timbers, Hamish Ivison, Jacob Morrison, Jake Poznanski, Kyle Lo, Luca Soldaini, Matt Jordan, Mayee Chen, Michael Noukhovitch, Nathan Lambert, Pete Walsh, Pradeep Dasigi, Robert Berry, Saumya Malik, Saurabh Shah, Scott Geng, Shane Arora, Shashank Gupta, Taira Anderson, Teng Xiao, Tyler Murray, Tyler Romero, Victoria Graf, Akari Asai, Akshita Bhagia, Alexander Wettig, Alisa Liu, Aman Rangapur, Chloe Anastasiades, Costa Huang, Dustin Schwenk, Harsh Trivedi, Ian Magnusson, Jaron Lochner, Jiacheng Liu, Lester James~V. Miranda, Maarten Sap, Malia Morgan, Michael Schmitz, Michal Guerquin, Michael Wilson, Regan Huff, Ronan~Le Bras, Rui Xin, Rulin Shao, Sam Skjonsberg, Shannon~Zejiang Shen, Shuyue~Stella Li, Tucker Wilde, Valentina Pyatkin, Will Merrill, Yapei Chang, Yuling Gu, Zhiyuan Zeng, Ashish Sabharwal, Luke Zettlemoyer, Pang~Wei Koh, Ali Farhadi, Noah~A. Smith, and Hannaneh
  Hajishirzi.
\newblock Olmo 3, 2026.
\newblock URL \url{https://arxiv.org/abs/2512.13961}.

\bibitem[Nadimi et~al.(2025)Nadimi, Boutaib, and Zheng]{verilog}
Bardia Nadimi, Ghali~Omar Boutaib, and Hao Zheng.
\newblock Pyranet: A multi-layered hierarchical dataset for verilog.
\newblock In \emph{2025 62nd ACM/IEEE Design Automation Conference (DAC)}, page 1–7. IEEE, 2025.
\newblock \doi{10.1109/dac63849.2025.11133406}.
\newblock URL \url{http://dx.doi.org/10.1109/DAC63849.2025.11133406}.

\bibitem[Hendrycks et~al.(2021)Hendrycks, Burns, Basart, Zou, Mazeika, Song, and Steinhardt]{mmlu}
Dan Hendrycks, Collin Burns, Steven Basart, Andy Zou, Mantas Mazeika, Dawn Song, and Jacob Steinhardt.
\newblock Measuring massive multitask language understanding, 2021.
\newblock URL \url{https://arxiv.org/abs/2009.03300}.

\bibitem[Talmor et~al.(2019)Talmor, Herzig, Lourie, and Berant]{commonsenseqa}
Alon Talmor, Jonathan Herzig, Nicholas Lourie, and Jonathan Berant.
\newblock Commonsenseqa: A question answering challenge targeting commonsense knowledge, 2019.
\newblock URL \url{https://arxiv.org/abs/1811.00937}.

\bibitem[Clark et~al.(2018)Clark, Cowhey, Etzioni, Khot, Sabharwal, Schoenick, and Tafjord]{arc}
Peter Clark, Isaac Cowhey, Oren Etzioni, Tushar Khot, Ashish Sabharwal, Carissa Schoenick, and Oyvind Tafjord.
\newblock Think you have solved question answering? try arc, the ai2 reasoning challenge, 2018.
\newblock URL \url{https://arxiv.org/abs/1803.05457}.

\bibitem[Aljundi et~al.(2018)Aljundi, Babiloni, Elhoseiny, Rohrbach, and Tuytelaars]{aljundi2018memory}
Rahaf Aljundi, Francesca Babiloni, Mohamed Elhoseiny, Marcus Rohrbach, and Tinne Tuytelaars.
\newblock Memory aware synapses: Learning what (not) to forget, 2018.
\newblock URL \url{https://arxiv.org/abs/1711.09601}.

\bibitem[Rebuffi et~al.(2017)Rebuffi, Kolesnikov, Sperl, and Lampert]{rebuffi2017icarlincrementalclassifierrepresentation}
Sylvestre-Alvise Rebuffi, Alexander Kolesnikov, Georg Sperl, and Christoph~H. Lampert.
\newblock icarl: Incremental classifier and representation learning, 2017.
\newblock URL \url{https://arxiv.org/abs/1611.07725}.

\bibitem[Shin et~al.(2017)Shin, Lee, Kim, and Kim]{shin2017continuallearningdeepgenerative}
Hanul Shin, Jung~Kwon Lee, Jaehong Kim, and Jiwon Kim.
\newblock Continual learning with deep generative replay, 2017.
\newblock URL \url{https://arxiv.org/abs/1705.08690}.

\bibitem[Adila et~al.(2026)Adila, Mazzawi, Dherin, and Gonzalvo]{adila2026grow}
Dyah Adila, Hanna Mazzawi, Benoit Dherin, and Xavier Gonzalvo.
\newblock Grow, don't overwrite: Fine-tuning without forgetting, 2026.
\newblock URL \url{https://arxiv.org/abs/2603.08647}.

\bibitem[Yan et~al.(2021)Yan, Xie, and He]{der}
Shipeng Yan, Jiangwei Xie, and Xuming He.
\newblock Der: Dynamically expandable representation for class incremental learning, 2021.
\newblock URL \url{https://arxiv.org/abs/2103.16788}.

\bibitem[Wang et~al.(2022)Wang, Zhou, Ye, and Zhan]{foster}
Fu-Yun Wang, Da-Wei Zhou, Han-Jia Ye, and De-Chuan Zhan.
\newblock Foster: Feature boosting and compression for class-incremental learning, 2022.
\newblock URL \url{https://arxiv.org/abs/2204.04662}.

\bibitem[Prabhu et~al.(2023)Prabhu, Hammoud, Dokania, Torr, Lim, Ghanem, and Bibi]{prabhu2023computationally}
Ameya Prabhu, Hasan Abed Al~Kader Hammoud, Puneet Dokania, Philip H.~S. Torr, Ser-Nam Lim, Bernard Ghanem, and Adel Bibi.
\newblock Computationally budgeted continual learning: What does matter?, 2023.
\newblock URL \url{https://arxiv.org/abs/2303.11165}.

\bibitem[Cho et~al.(2026)Cho, Moon, Chunara, Cho, and Cha]{cho2026forgetforgettingcontinuallearning}
Dongkyu Cho, Taesup Moon, Rumi Chunara, Kyunghyun Cho, and Sungmin Cha.
\newblock Forget forgetting: Continual learning in a world of abundant memory, 2026.
\newblock URL \url{https://arxiv.org/abs/2502.07274}.

\bibitem[Dohare et~al.(2024)Dohare, Hernandez-Garcia, Lan, Rahman, Mahmood, and Sutton]{dohare2024loss}
Shibhansh Dohare, J~Fernando Hernandez-Garcia, Qingfeng Lan, Parash Rahman, A~Rupam Mahmood, and Richard~S Sutton.
\newblock Loss of plasticity in deep continual learning.
\newblock \emph{Nature}, 632\penalty0 (8026):\penalty0 768--774, 2024.
\newblock URL \url{https://www.nature.com/articles/s41586-024-07711-7}.

\bibitem[Mitchell et~al.(2025)Mitchell, Alliegro, Camoriano, Carrión-Ojeda, Carta, Chalvatzaki, Churamani, D'Eramo, Hamidi, Hesse, Hinder, Kamath, Lomonaco, Paul, Pistilli, Tuytelaars, van~de Ven, Kersting, Schaub-Meyer, and Mundt]{mitchell2025continual}
Rupert Mitchell, Antonio Alliegro, Raffaello Camoriano, Dustin Carrión-Ojeda, Antonio Carta, Georgia Chalvatzaki, Nikhil Churamani, Carlo D'Eramo, Samin Hamidi, Robin Hesse, Fabian Hinder, Roshni~Ramanna Kamath, Vincenzo Lomonaco, Subarnaduti Paul, Francesca Pistilli, Tinne Tuytelaars, Gido~M van~de Ven, Kristian Kersting, Simone Schaub-Meyer, and Martin Mundt.
\newblock Continual learning should move beyond incremental classification, 2025.
\newblock URL \url{https://arxiv.org/abs/2502.11927}.

\bibitem[Wu et~al.(2024)Wu, Luo, Li, Pan, Vu, and Haffari]{wu2024continuallearninglargelanguage}
Tongtong Wu, Linhao Luo, Yuan-Fang Li, Shirui Pan, Thuy-Trang Vu, and Gholamreza Haffari.
\newblock Continual learning for large language models: A survey, 2024.
\newblock URL \url{https://arxiv.org/abs/2402.01364}.

\bibitem[Cho et~al.(2025)Cho, Hwang, and Lee]{cho2025peer}
Dong~Kyu Cho, Inwoo Hwang, and Sanghack Lee.
\newblock Peer pressure: Model-to-model regularization for single source domain generalization, 2025.
\newblock URL \url{https://arxiv.org/abs/2505.12745}.

\bibitem[Kirichenko et~al.(2021)Kirichenko, Farajtabar, Rao, Lakshminarayanan, Levine, Li, Hu, Wilson, and Pascanu]{hcl}
Polina Kirichenko, Mehrdad Farajtabar, Dushyant Rao, Balaji Lakshminarayanan, Nir Levine, Ang Li, Huiyi Hu, Andrew~Gordon Wilson, and Razvan Pascanu.
\newblock Task-agnostic continual learning with hybrid probabilistic models.
\newblock In \emph{ICML Workshop on Invertible Neural Networks, Normalizing Flows, and Explicit Likelihood Models}, 2021.
\newblock URL \url{https://openreview.net/forum?id=ZbSeZKdqNkm}.

\bibitem[Resta and Bacciu(2024)]{resta2024selfgeneratedreplaymemoriescontinual}
Michele Resta and Davide Bacciu.
\newblock Self-generated replay memories for continual neural machine translation, 2024.
\newblock URL \url{https://arxiv.org/abs/2403.13130}.

\bibitem[Ke et~al.(2023)Ke, Shao, Lin, Konishi, Kim, and Liu]{ke2023continualpretraininglanguagemodels}
Zixuan Ke, Yijia Shao, Haowei Lin, Tatsuya Konishi, Gyuhak Kim, and Bing Liu.
\newblock Continual pre-training of language models, 2023.
\newblock URL \url{https://arxiv.org/abs/2302.03241}.

\bibitem[Baek et~al.(2026)Baek, Monti, Schwab, Abbas, Adiga, Blakeney, Böther, Burstein, Carranza, Deng, Doshi, Dorna, Fang, Jiang, Joshi, Larsen, Lee, Mentzer, Merrick, Mongstad, Pan, Suri, Teh, Telanoff, Urbanek, Wang, Wills, Yin, Raghunathan, Kolter, Gaza, Morcos, Leavitt, and Maini]{baek2026finetunersfallacypretrainfinetuning}
Christina Baek, Ricardo~Pio Monti, David Schwab, Amro Abbas, Rishabh Adiga, Cody Blakeney, Maximilian Böther, Paul Burstein, Aldo~Gael Carranza, Alvin Deng, Parth Doshi, Vineeth Dorna, Alex Fang, Tony Jiang, Siddharth Joshi, Brett~W. Larsen, Jason~Chan Lee, Katherine~L. Mentzer, Luke Merrick, Haakon Mongstad, Fan Pan, Anshuman Suri, Darren Teh, Jason Telanoff, Jack Urbanek, Zhengping Wang, Josh Wills, Haoli Yin, Aditi Raghunathan, J.~Zico Kolter, Bogdan Gaza, Ari Morcos, Matthew Leavitt, and Pratyush Maini.
\newblock The finetuner's fallacy: When to pretrain with your finetuning data, 2026.
\newblock URL \url{https://arxiv.org/abs/2603.16177}.

\bibitem[Korbak et~al.(2023)Korbak, Shi, Chen, Bhalerao, Buckley, Phang, Bowman, and Perez]{korbak2023pretraininglanguagemodelshuman}
Tomasz Korbak, Kejian Shi, Angelica Chen, Rasika Bhalerao, Christopher~L. Buckley, Jason Phang, Samuel~R. Bowman, and Ethan Perez.
\newblock Pretraining language models with human preferences, 2023.
\newblock URL \url{https://arxiv.org/abs/2302.08582}.

\bibitem[Springer et~al.(2025)Springer, Goyal, Wen, Kumar, Yue, Malladi, Neubig, and Raghunathan]{springer2025overtrainedlanguagemodelsharder}
Jacob~Mitchell Springer, Sachin Goyal, Kaiyue Wen, Tanishq Kumar, Xiang Yue, Sadhika Malladi, Graham Neubig, and Aditi Raghunathan.
\newblock Overtrained language models are harder to fine-tune, 2025.
\newblock URL \url{https://arxiv.org/abs/2503.19206}.

\bibitem[Liu et~al.(2022)Liu, Xie, Li, and Ma]{liu2022pretraininglossbetterdownstream}
Hong Liu, Sang~Michael Xie, Zhiyuan Li, and Tengyu Ma.
\newblock Same pre-training loss, better downstream: Implicit bias matters for language models, 2022.
\newblock URL \url{https://arxiv.org/abs/2210.14199}.

\bibitem[Watts et~al.(2026)Watts, Li, Goyal, Springer, and Raghunathan]{sharpness_aware_pretraining}
Ishaan Watts, Catherine Li, Sachin Goyal, Jacob~Mitchell Springer, and Aditi Raghunathan.
\newblock Sharpness-aware pretraining mitigates catastrophic forgetting.
\newblock \emph{arXiv preprint arXiv:2605.02105}, 2026.

\bibitem[Kim et~al.(2025)Kim, Lee, Cho, Jang, Hwang, Won, Ahn, Lee, and Seo]{kim2024knowledge}
Jiyeon Kim, Hyunji Lee, Hyowon Cho, Joel Jang, Hyeonbin Hwang, Seungpil Won, Youbin Ahn, Dohaeng Lee, and Minjoon Seo.
\newblock Knowledge entropy decay during language model pretraining hinders new knowledge acquisition, 2025.
\newblock URL \url{https://arxiv.org/abs/2410.01380}.

\bibitem[Su et~al.(2024)Su, Ahmed, Lu, Pan, Bo, and Liu]{rope}
Jianlin Su, Murtadha Ahmed, Yu~Lu, Shengfeng Pan, Wen Bo, and Yunfeng Liu.
\newblock Roformer: Enhanced transformer with rotary position embedding.
\newblock \emph{Neurocomputing}, 568:\penalty0 127063, 2024.

\bibitem[Zhang and Sennrich(2019)]{rmsnorm}
Biao Zhang and Rico Sennrich.
\newblock Root mean square layer normalization.
\newblock \emph{Advances in Neural Information Processing Systems}, 32, 2019.

\bibitem[Hendrycks and Gimpel(2016)]{gelu}
Dan Hendrycks and Kevin Gimpel.
\newblock Gaussian error linear units (gelus).
\newblock \emph{arXiv preprint arXiv:1606.08415}, 2016.

\bibitem[Loshchilov and Hutter(2019)]{loshchilov2019decoupledweightdecayregularization}
Ilya Loshchilov and Frank Hutter.
\newblock Decoupled weight decay regularization, 2019.
\newblock URL \url{https://arxiv.org/abs/1711.05101}.

\bibitem[Jouppi et~al.(2023)Jouppi, Kurian, Li, Ma, Nagarajan, Nai, Patil, Subramanian, Swing, Towles, et~al.]{tpuv4}
Norm Jouppi, George Kurian, Sheng Li, Peter Ma, Rahul Nagarajan, Lifeng Nai, Nishant Patil, Suvinay Subramanian, Andy Swing, Brian Towles, et~al.
\newblock Tpu v4: An optically reconfigurable supercomputer for machine learning with hardware support for embeddings.
\newblock In \emph{Proceedings of the 50th annual international symposium on computer architecture}, pages 1--14, 2023.

\bibitem[Bradbury et~al.(2018)Bradbury, Frostig, Hawkins, Johnson, Leary, Maclaurin, Necula, Paszke, Vander{P}las, Wanderman-{M}ilne, and Zhang]{jax}
James Bradbury, Roy Frostig, Peter Hawkins, Matthew~James Johnson, Chris Leary, Dougal Maclaurin, George Necula, Adam Paszke, Jake Vander{P}las, Skye Wanderman-{M}ilne, and Qiao Zhang.
\newblock {JAX}: composable transformations of {P}ython+{N}um{P}y programs, 2018.
\newblock URL \url{http://github.com/jax-ml/jax}.

\end{thebibliography}

\clearpage
\appendix

\section{Method Details}
\label[appendix]{app:method-details}

\subsection{Self-generated Replay.}

We generate replay data from the frozen reference model. Each sequence starts with a corpus identifier, functionally similar to a Beginning-of-Sequence (BOS) token but unique to each dataset. If we had instead started each sequence with a standard BOS token, the model wouldn't know which distribution to sample the second token from -- hence the measured pretraining loss would go up, even if the model did not forget any information. The remaining tokens are sampled autoregressively from the reference model at temperature $1$ until a fixed sequence length is reached.

\section{Experimental Details}
\label[appendix]{app:exp-details}

Each experiment is fully specified in our codebase, from training scripts and run configurations to plotting scripts. Below we provide a high-level summary; for low-level details, please refer to the codebase: \url{https://github.com/martin-marek/forgetting}.

\subsection{Model architecture}

We use the following transformer-decoder language model architecture for every experiment except \Cref{fig:instruct}. The architecture uses RoPE \citep{rope}, RMSNorm \citep{rmsnorm}, GELU \citep{gelu}, and untied embeddings.

\par\noindent
\begin{minipage}{\linewidth}
\begin{lstlisting}[style=paperpython]
def forward(tokens, weights):
   h = weights['embed_in'][tokens]
   for w in weights['layers']:
       q, k, v = jnp.einsum('btd,sndh->sbtnh', rms_norm(h), w['qkv'])
       q, k = apply_rope(rms_norm(q)), apply_rope(rms_norm(k))
       a = jax.nn.dot_product_attention(q, k, v, is_causal=True)
       h += jnp.einsum('btnh,nhd->btd', a, w['out'])
       m = jax.nn.gelu(jnp.einsum('btd,df->btf', rms_norm(h), w['up']))
       h += jnp.einsum('btf,fd->btd', m, w['down'])
   return jnp.einsum('btd,vd->btv', rms_norm(h), weights['embed_out'])
\end{lstlisting}
\end{minipage}

\subsection{Implementation Details.}

Except for \Cref{fig:instruct}, every experiment uses batch size $B=256$, context length $T=256$, and AdamW~\citep{loshchilov2019decoupledweightdecayregularization} with $(\beta_1,\beta_2)=(0.9,0.999)$ and weight decay $0.02$, with a linear learning rate warmup. We generally use cosine learning rate decay for both pretraining and finetuning except for runs where the training duration cannot be statically determined, in particular: \Cref{fig:learning_rate,fig:capacity_detailed} where we train until a fixed target loss, and \Cref{fig:instruct,fig_method_comparison} where we train until the validation loss starts to increase.

The replay regularization is computed on minibatches drawn from the pretraining distribution, using by default a batch size 4-times smaller than the downstream batch (to reduce the computational overhead from regularization).

\subsection{Compute Resources}
\label[appendix]{appendix:compute_resources}

We run our experiments on Google TPU v6e-8 VMs \citep{tpuv4} in JAX \citep{jax}. Larger sweeps are run on multiple (up to 16) v6e-8 workers in parallel, with each worker running an independent pretraining or finetuning job.

\end{document}